\documentclass{article}
\PassOptionsToPackage{numbers}{natbib}
\usepackage[preprint]{neurips_2019}
\usepackage[utf8]{inputenc} % allow utf-8 input
\usepackage[T1]{fontenc}    % use 8-bit T1 fonts
\usepackage{hyperref}       % hyperlinks
\usepackage{url}            % simple URL typesetting
\usepackage{booktabs}       % professional-quality tables
\usepackage{amsfonts}       % blackboard math symbols
\usepackage{nicefrac}       % compact symbols for 1/2, etc.
\usepackage{microtype}      % microtypography
\usepackage{subcaption}
\usepackage{graphicx}
\usepackage{amsthm}
\usepackage{algorithm}
\usepackage{enumitem}
\usepackage{algpseudocode}
\usepackage[normalem]{ulem}
\usepackage[table,xcdraw]{xcolor}
\usepackage{units}
\usepackage{appendix}
\newtheorem{theorem}{Theorem}[section]
\newtheorem{definition}{Definition}[section]
\newenvironment{proofsketch}{%
\proof}{\endproof}
\def \Rtwo{\textbf{a}}
\def \Rthree{\textbf{b}}
\def \Rfour{\textbf{c}}
\def \Rfive{\textbf{d}}
\def \DNconn{\textbf{a}}
\def \DNpath{\textbf{b}}
\def \PPvar{\textbf{a}}

\title{5 Parallel Prism: A topology for pipelined implementations of
convolutional neural networks using computational memory}

\author{%
Martino Dazzi\textsuperscript{1, 2}, Abu Sebastian\textsuperscript{1}, Pier Andrea Francese\textsuperscript{1}, \\\textbf{Thomas Parnell\textsuperscript{1}, Luca Benini\textsuperscript{2} and Evangelos Eleftheriou\textsuperscript{1}}\\
\textsuperscript{1} IBM Research -- Zurich, Switzerland,  \textsuperscript{2} ETH Zurich, Switzerland\\
\texttt{ \{daz, ase, pfr,  tpa, ele\}@zurich.ibm.com}\\ \texttt{lbenini@iis.ee.ethz.ch} \\
}

\begin{document}
\bibliographystyle{abbrvnat}
% \nipsfinalcopy is no longer used

\maketitle
\begin{abstract}
In-memory computing is an emerging computing paradigm that could enable deep-learning inference at significantly higher energy efficiency and reduced
latency. The essential idea is to map the synaptic weights corresponding to each layer to one or more computational memory (CM) cores. During
inference, these cores perform the associated matrix-vector multiply operations in place with O(1) time complexity, thus obviating the need to move
the synaptic weights to an additional processing unit. Moreover, this architecture could enable the execution of these networks in a highly pipelined
fashion. However, a key challenge is to design an efficient communication fabric for the CM cores. Here, we present one such communication
fabric based on a graph topology that is well suited for the widely successful convolutional neural networks (CNNs). We show that this
communication fabric facilitates the pipelined execution of all state-of-the-art CNNs by proving the existence of a homomorphism between one graph
representation of these networks and the proposed graph topology. We then present a quantitative comparison with established communication topologies
and show that our proposed topology achieves the lowest bandwidth requirements per communication channel. Finally, we present a concrete example of
mapping ResNet-32 onto an array of CM cores.
\end{abstract}

\section{Introduction} \label{sec:intro}
Deep neural networks (DNN) have revolutionized the field of machine learning by providing unprecedented human-like performance in solving many
real-world problems such as image and speech recognition. However, the training and inference of large DNNs is a computationally intensive task and
has motivated the search for novel computing architectures targeting this application~\citep{Y2018fleischerVLSI, Y2017luHPCA, Y2011farabetCVPR, Y2018wangNIPS}. Recent years have seen an explosion
of companies developing customized hardware accelerators for DNN training and inference. Companies are motivated to develop such
hardware both to accelerate their large-scale internal DNN workloads and to give access to such devices on a pay-per-use basis in the
cloud. The DAWNBench benchmark for training time on ImageNet is currently topped by a system using GPUs with specialized cores designed for DNN
training, and the corresponding benchmark for training cost is topped by a cloud-based workload running on TPUs: a customized chip specifically
designed for DNN training and inference~\citep{Y2017jouppiISCA}.

Although customized accelerators are now considered mainstream in the machine-learning community, they still suffer from the inherent
limitations of von Neumann computing architectures. Namely, synaptic weights must be repeatedly moved between the memory units and the compute units.
This bottleneck leads many to consider alternative non-von Neumann architectures such as in-memory computing~\citep{Y2017burrAPX,Y2018legalloNatureElectronics,Y2018huAdvMat,Y2018ielminiNatureElectronics,Y2015preziosoNature,Y2019xiaNature}. Fundamentally, by taking advantage of a set of memristive devices organized in a crossbar array, one can leverage the physical properties of these devices via Ohm's and Kirchhoff's laws to perform in-place matrix-vector multiplication with $O(1)$ time complexity. As DNN training and inference is dominated by such operations, which
require $O(N^2)$ time with traditional architectures, the potential for acceleration is self-evident. It was also shown recently that it is possible
to achieve near software accuracies when synaptic weights are stored in phase-change memory devices fabricated in the \unit[90]~{nm} technology node~\citep{Y2019sebastianVLSI}.

However, the ability to perform matrix-vector multiplication in constant time, as well as the one-to-one mapping between synaptic weights and
memristive devices, creates a new challenge for  designers of such accelerators. If we want to perform computation in a
\textit{pipelined} manner, the bottleneck becomes how quickly we can communicate activations from one layer/computational memory (CM) core to
another. This clearly depends on both the architecture of the neural network model itself, i.e.~how the different layers are connected, as well as
the topology of the communication fabric of the accelerator device, i.e.~how long it takes to transfer activations between physically distinct
layers. To manufacture a general-purpose inference engine, the
challenge would be to design a communication topology that supports low-latency
execution for a wide range of neural network architectures without introducing unnecessary
redundancy, e.g.~unused~connections.

\textbf{Contributions} We attempt to address exactly the above problem with a focus on convolutional neural networks (CNN)
architectures due to their popularity in a range of tasks and their suitability for in-memory computing: Every layer executes with $O(1)$ time complexity irrespective of channel depth, and the computation of activations can be pipelined across different layers, making full use of the physically instantiated neurons. Specifically, our contributions are:
\begin{itemize}
\item We propose a novel graph topology (5 Parallel Prism, or 5PP) for pipelined executions of CNNs that maximally exploits the physical proximity of
computational units. 
\item We prove theoretically that all state-of-the-art neural networks featuring feedforward, residual, Inception-style and
dense connectivity are mappable onto the proposed topology.
\item We present a quantitative comparison of 5PP with existing communication topologies
and demonstrate its advantages in terms of latency, throughput and efficiency. 
\item We give a detailed example of how to map ResNet-32 onto an
array of CM elements.
\end{itemize}

\section{A Consolidated Graph Representation of CNNs}\label{sec:cnngraphs}
\begin{figure}[h!]
\centering
\includegraphics[width=1\linewidth]{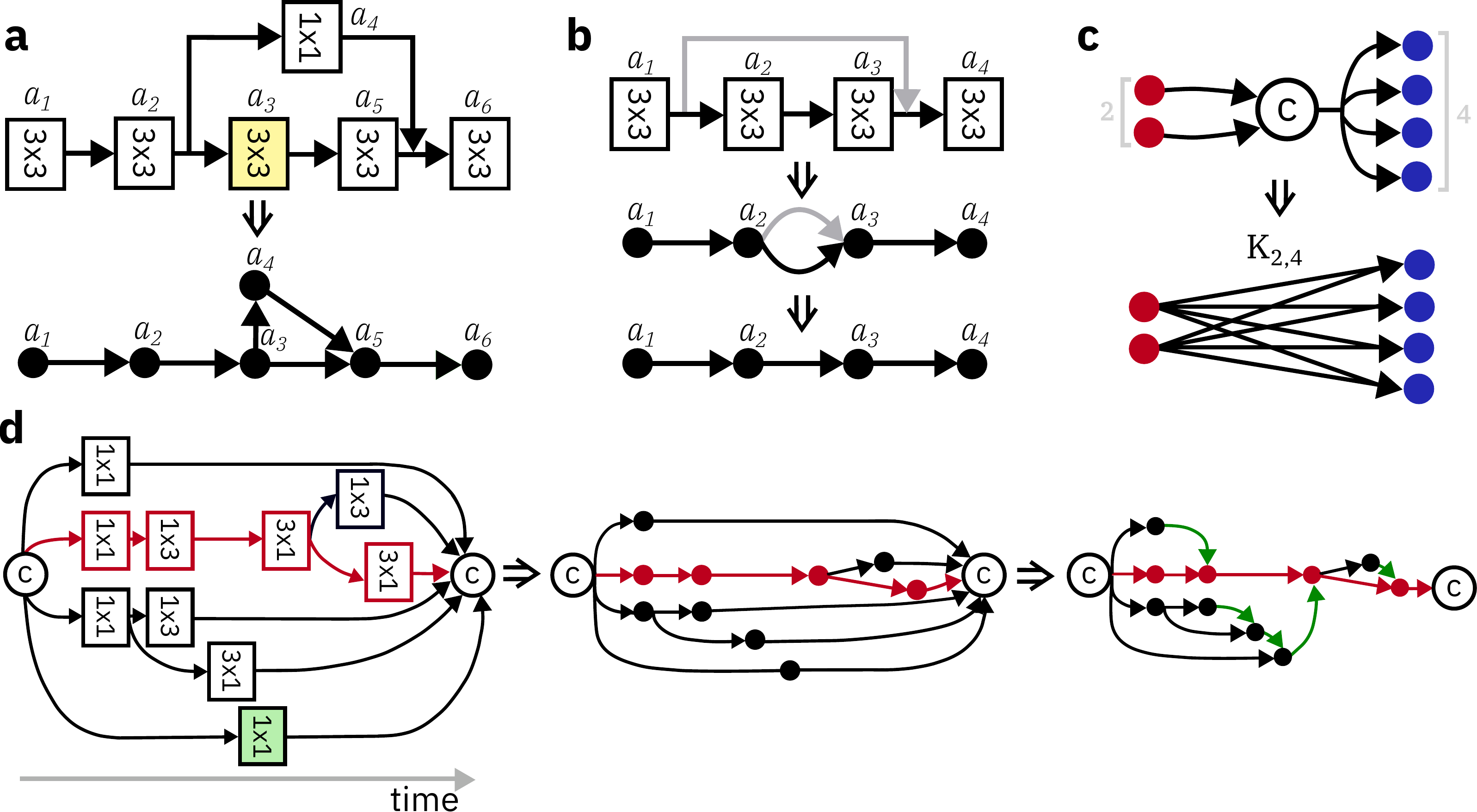}
\caption{Illustration of representation rules. Yellow layer in \textbf{a.} represents a layer with stride = 2; green layer in \textbf{d.} represents a layers to which 3$\times$3 max pooling is previously applied.}
%\caption{\textbf{a.} Rule 2 applied to the resampling layer of ResNet. S indicates the stride. In our representation the residual activations are collected from the input memory of $a_{2}$, and added to the output memory of $a_{5}$. \textbf{b.} Rule 3 applied to the residual connection of ResNet. The presence of a residual connection has to be taken into account when designing the bandwidth of the physical link, but is redundant in terms of graph representation. \textbf{c.} Inception-C block from the Inception v4 architecture, with its layers organized by their time of activation processing in a pipeline. Yellow box indicates convolution after 3x3 max pooling operation. In the time axis, $T=1$ unit computational time, $ \tau = H \cdot T $, with H the height of the image. \textbf{d.} Application of rule 4 to the Inception-C block. Vertices that execute before the longest latency path has completed can hop their activation on other vertices. \textbf{e.} Illustration of rule 5. Assuming the concatenation operation is performed at the destination nodes, we can represent it equivalently with a complete bipartite graph, in the example $ K_{2,4} $.}
\label{Rules}
\end{figure}
A CNN can be implemented in CM in a pipelined fashion if there exists communication channels that match the connectivity of the
network. If this was not verified, the activations from one CM core may require transferring for several cycles of the execution schedule through different cores before they can reach their destination. As the execution happens in a pipelined fashion, every core would then systematically stall at every cycle until the slowest data transfer is completed, severely impairing the throughput. Mathematically:
\begin{definition} Given a communication fabric with topology $\mathcal{F}$ and the directed graph representation
$\mathcal{C}$ of a CNN, with vertices representing convolutional layers and edges representing activations directed toward the direction of computation,
then the CNN is executable in a pipelined fashion on $\mathcal{F}$ if there exists a homomorphism $ h:\mathcal{C} \rightarrow \mathcal{F}$.
\end{definition}
As the communication fabric will have to provide enough connection overhead to accommodate a diverse set of networks, the homomorphism will generally
be injective. The definition of a homomorphism between a directed and an undirected graph is justified by the fact that, in our representation of the
communication fabric, undirected edges represent bi-directional communication channels, which can be assigned either direction when the CNN is mapped
onto it.

Here we present one such graph representation of CNNs, $\mathcal{C}$.
Typically, CNN architectures are represented as directed graphs, with vertices representing convolutional
layers and directed edges representing activations directed toward their next layers. Nevertheless,
CNN representations are not coherent with their physical implementation, some examples
being input images or concatenation operations themselves being represented as one vertex of the
network, or pooling operations being at times represented along convolutions in the same vertex and
at times separately. Based on these consolidations, we present five design rules:
\begin{enumerate}[leftmargin=2em]
\item[R1.] Vertices are identified solely with convolutional layers. Any other operation, such as pooling or addition for the residual path, is
treated as pre- or post-processing of the convolution.
%As the multiply-and-accumulate operation of the convolutional layers is the most computationally intensive, this makes sense also for any digital pipeline implementation.
%\item[R2.] Although standard CNN representations are in fact graphs, they are not devoid of the concept of input and output of the layer. Fig. \ref{Rules}\Rtwo, showing the residual path across layers with different strides from the ResNet architecture, is taken as an example. In the classic representation, the left side of the boxes stands for the input of the layers and the right side its output: in the residual path, $a_{4} $‘s output adds to the output of $a_{5}$. This is problematic in a graph representation, where there is no concept of input and output of a vertex. Physically, the distinction between the input and output of the layer translates to the distinction between the input memory (the operands) and output memory (the result) in the computational unit. In our representation, we abstract from the concept of input and output memory, meaning the communication links between computational units see a unified memory interface. This abstraction rids the graph from the idea of input and output of the vertex, as can be seen in the bottom half of Fig. \ref{Rules}\Rtwo, where now  $a_{4}$ simply connects to $a_{5}$.
\item[R2.] We do not distinguish between input and output of a vertex. Vertices are either connected or not connected. This is illustrated in Fig.~\ref{Rules}\Rtwo.
\item[R3.] Edges that may make the graph non-simple are removed. This is illustrated for the ResNet architecture in Fig.~\ref{Rules}\Rthree.
%One example is the ResNet architecture(Fig. \ref{Rules}\Rthree), in which, once unified the ideas of input and output of the layer, every residual connection would be represented by an edge parallel to the feedforward connection. In our representation, graphs are simplified and thus parallel edges are removed.
\item[R4.] Concatenation does not imply any operation on the data, thus it cannot be represented as a vertex in the graph. Given this assumption, the
concatenation of \textit{m} vertices being fed into \textit{n} others is equivalent to a complete bipartite graph $k_{m,n}$ as in Fig.~\ref{Rules}\Rfour.
\item[R5.] A series-parallel (s-p) graph such as Inception can have some parallel path with latency much smaller than the critical path (see Fig.~\ref{Rules}\Rfive, with the nodes laid out in order of latency and the critical path in red) that are unable to reach their physical destination unless they \textit{hop} to the neighboring vertex as the critical path is
executing. We require that such activations hop to their nearest temporally subsequent node (green arrows in Fig.~\ref{Rules}\Rfive).

\end{enumerate}

\section{A New Communication Fabric: 5 Parallel Prism}\label{sec:5PP}

\begin{figure}[h!]
\centering
\includegraphics[width=.77\linewidth]{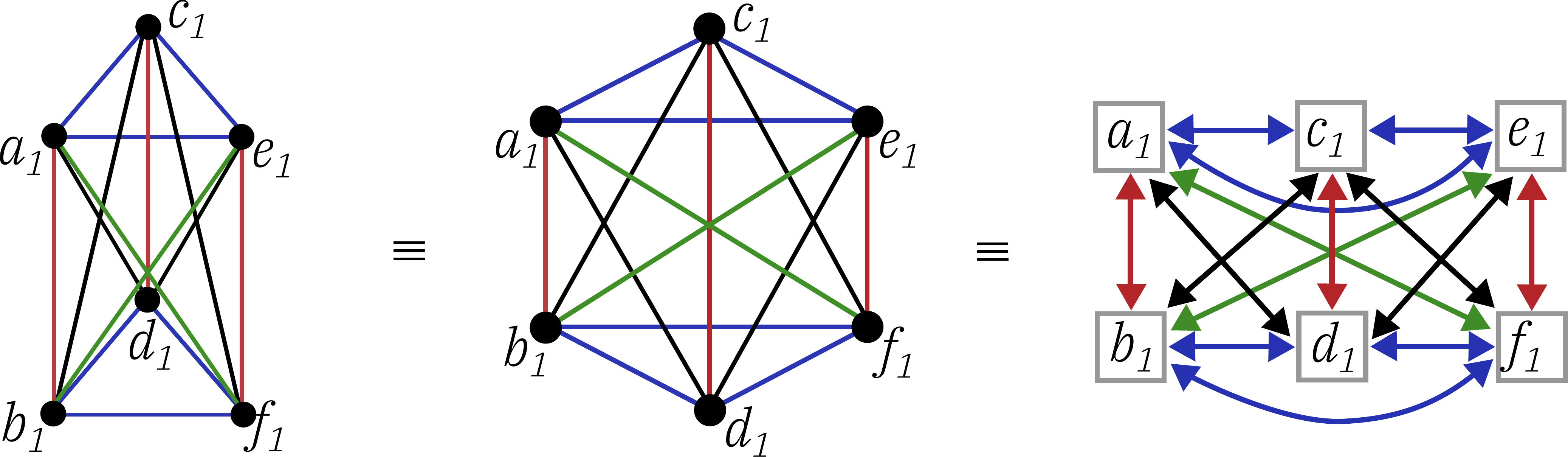}
\caption{Unit graph, from left to right: our 3D representation, its canonical graph theory representation as a complete graph $K_{6}$, and its physical
implementation, with vertices representing computational units and edges representing bidirectional links.} \label{unit_graph}
\end{figure}

\begin{algorithm}[t]
\caption{5 Parallel Prism Construction}\label{alg:5pp_construction}
\begin{algorithmic}[1]
\State $\textit{ \% The initial disjoint union of unit graphs has vertices: } \{ \tilde{a}_{1}, \tilde{b}_{1}, ..., \tilde{e}_{1}, \tilde{f}_{1}, ..., \tilde{a}_{M}, ..., \tilde{f}_{M} \}$
\State $\textit{ \% It creates a 5PP with vertices } \{ a_{1}, ..., a_{M}, b_{1}, ..., b_{M}, c_{M},d_{M},e_{M},f_{M} \} $
\State $\textit{ \% The first unit graph comprises the first 6 vertices of the 5PP } $
\State
\State $ \{ a_{1}, b_{1}, c_{1}, d_{1}, e_{1}, f_{1}  \}= \{ \tilde{a_{1}},\tilde{b_{1}}, \tilde{c_{1}}, \tilde{d_{1}}, \tilde{e_{1}}, \tilde{f_{1}}  \};$
\For {$j = 1:(M-1)$}
\State $a_{j+1}=c_{j}\odot \tilde{a}_{j+1};$
\State $b_{j+1}=d_{j}\odot \tilde{b}_{j+1};$
\State $c_{j+1}=e_{j}\odot \tilde{c}_{j+1};$
\State $d_{j+1}=f_{j}\odot \tilde{d}_{j+1};$
\State $e_{j+1}=\tilde{e}_{j+1};$
\State $f_{j+1}=\tilde{f}_{j+1};$
\EndFor
\end{algorithmic}
\end{algorithm}

%In this section we present our proposed topology. The topology is represented as an undirected graph, and is a  mathematical abstraction of a physical communication fabric implemented on a 2D array of computational units. In the graph representation of the communication fabric we present hereafter, we identify the computational units with the vertices and bidirectional communication channels with the undirected edges.

\begin{figure}[h!]
\centering
\includegraphics[width=.55\linewidth]{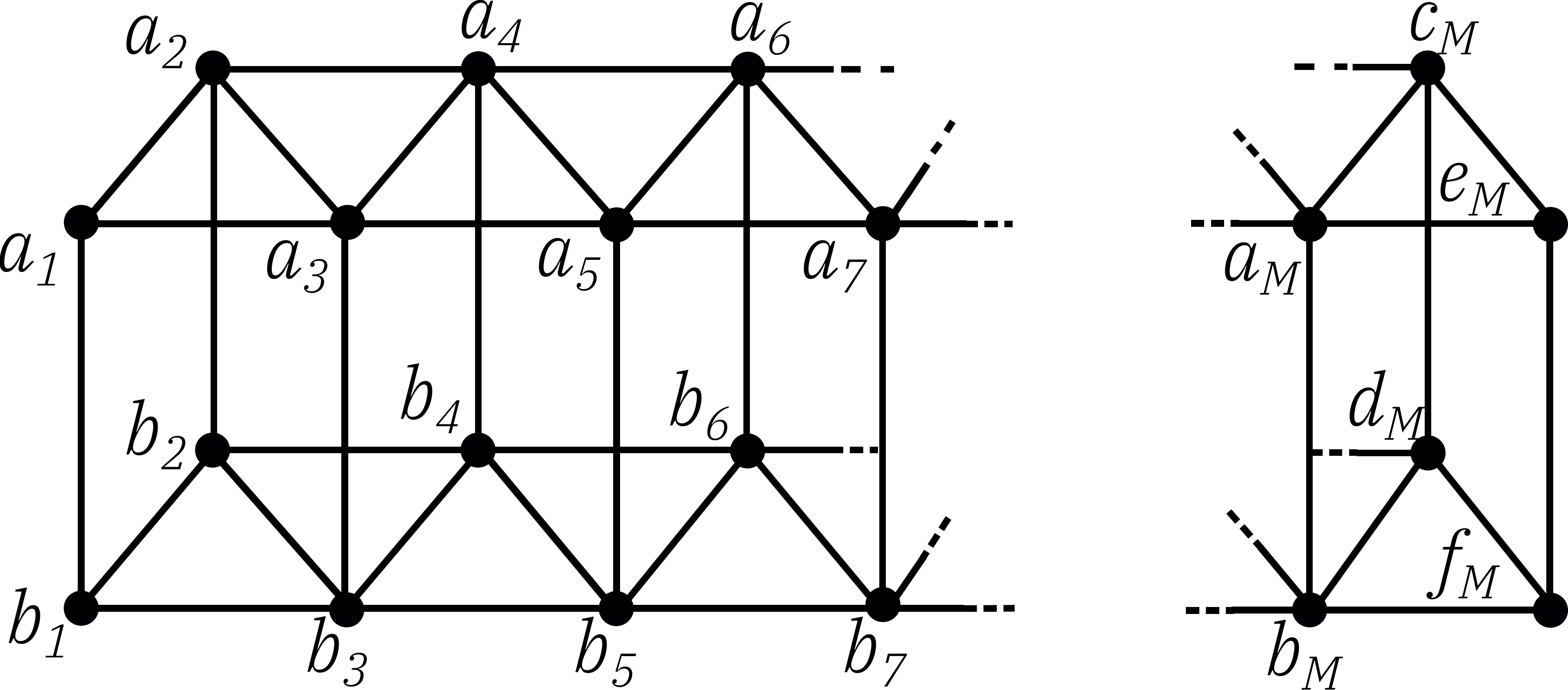}
\caption{5 Parallel Prism. The pseudocode in Algorithm~\ref{alg:5pp_construction} produces a 5PP with $N$ vertices $ \{ a_{1}, a_{2}, ..., a_{M},b_{1},
b_{2}, ..., b_{M},c_{M}, d_{M}, e_{M}, f_{M}  \}$; $M=\lceil \left( n-6 \right) / 2\rceil+1$. All diagonal edges of the prisms are omitted for clarity.} \label{fig:complete_prism}
\end{figure}

Here we propose a topology $\mathcal{F}$ that can be thought of as a graph spawning from the coalesce of multiple smaller graphs, referred to as ``unit graphs''. Figure~\ref{unit_graph} portrays the unit graph as an undirected complete graph $ K_{6}$. The physical implementation of the unit graph in Fig.~\ref{unit_graph}
clearly depicts how we maximize the use of spacial proximity in the interconnection of CM units with non-negligible
physical size: The basic communication infrastructure takes place between a 2-by-3 neighborhood, which is the way of packing six computational units on
a meshgrid that yields the shortest intra-unit maximum distance.
%Also not that the physical unit graph has 2D reflectional symmetry, , which is desirable in therms of implementation.\\
The construction of an $N$-vertice 5PP  starts from a disjoint union of $M=\lceil \left( n-6 \right) / 2\rceil+1$ unit graphs and is described by a pseudocode in Algorithm~\ref{alg:5pp_construction}, with the resulting topology depicted in Fig.~\ref{fig:complete_prism}. To represent the overall
topology, all diagonal edges of the prisms are omitted for clarity. The construction uses a vertex identification operation
that keeps the graph simple as described in Eq.~\ref{eq:vertex_identification}, where $u \cdot v$ is the vertex identification operation, \textit{G} is the graph
to which it is applied and \textit{V(G)}, \textit{E(G)}, the set of its vertices and edges.
\begin{equation} \label{eq:vertex_identification}
u \odot v =  \left \{ u \cdot v \mid u,v \in V(G),  \exists \textrm{ at most 1 } (u \cdot v, x) \in E(G), \forall x \in V(G)   \right \} .
\end{equation}

\section{Mapping CNN architectures onto 5PP} \label{sec:5PPmap}
Here we establish the existence a homomorphism between the consolidated graph representation of four different
state-of-the-art CNN architectures and the 5PP topology.
The process of verifying the existence of the homomorphism is often referred to as \textit{H}-coloring~\citep{Y1990hellJCT}. We propose an
\textit{iterative H-coloring} method that verifies the \textit{H}-coloring of the vertices of the directed graph in their order from initial to final.
That is, at the $i$-th iteration of verifying the \textit{H}-coloring, vertex $v_{i}$ in graph $\mathcal{C}$ is colored on a vertex in graph $\mathcal{F}$ such that
there are enough edges to connect it to whatever previous vertices $v_{0,...,i-1}$ it is connected to. The homomorphism is proven if all vertices of $\mathcal{C}$ have been successfully \textit{H}-colored onto $\mathcal{F}$  at the last
iteration.
Generally speaking, the proof of existence of a homomorphism between two arbitrary graphs is an NP-hard problem for any non-bipartite graph.
Leveraging the regularity of graphs representing CNNs, we present some properties of the 5PP that facilitate verification of the existence of a
homomorphism with the topologies of state-of-the-art CNNs.

\begin{figure}[h!]
\centering
\includegraphics[width=.75\linewidth]{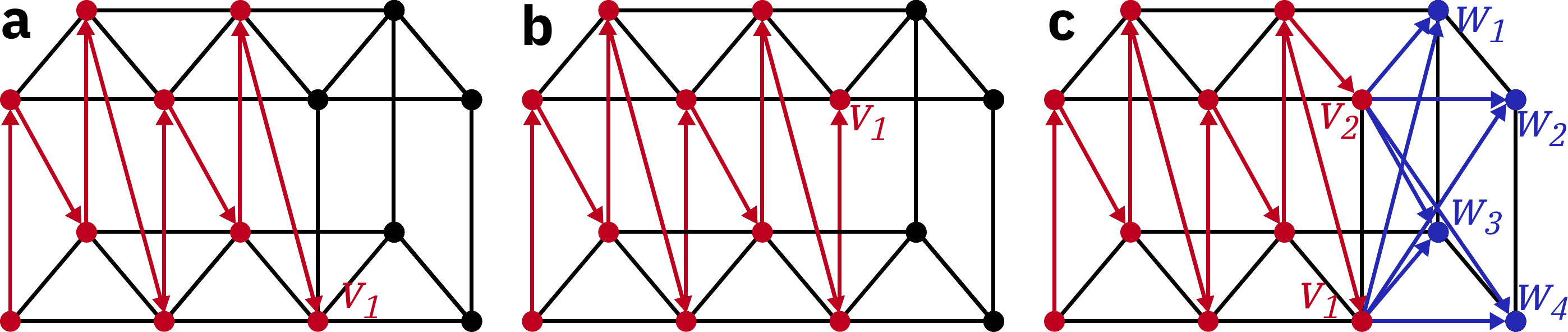}
\caption{\textbf{a.} Example of  \textit{H}-coloring of an odd-vertices path. Vertex $v_{1}$ can have maximum $d_{o} = 5$. \textbf{b.} Example of \textit{H}-coloring of an even-vertices path. Vertex $v_{1}$ can  have maximum $d_{o} = 4$. \textbf{c.} Example of \textit{H}-coloring of an even-vertices
tree. \textit{H}-coloring is then continued with complete bipartite graph $K_{2,4}$ between vertices $v_{1\sim2}$ and $w_{1\sim4}$.} \label{fig:P1P2}
\end{figure}
We define an \textit{even(odd)} \textit{H}-coloring as an \textit{H}-coloring that colors an even(odd) number of vertices. Furthermore, we define the out-degree $d^{o}$  of a
vertex \textit{v} of a directed graph as the number of outgoing edges from \textit{v}. Figure~\ref{fig:P1P2} illustrates three \textit{H}-colorings that are representative of the
two main properties of the 5PP:
\begin{enumerate}[leftmargin=2em]
\item[P1.] By construction, every vertex belongs to at least one complete graph $K_{6}$.
Given an odd(even) \textit{H}-coloring of the 5PP $ \{ a_{1},  b_{1}, ..., a_{n}, b_{n}, ...,a_{N} \}, \forall n \leq N$, vertex $a_{N}$ has possible maximum $ d^{o}= $5(4). The maximum number of vertices accessible in parallel in our topology gives it its name.
\item[P2.] Given any \textit{H}-coloring of the 5PP $ \{ a_{1},  b_{1}, ..., a_{n}, b_{n}, ...,a_{N} \}, \forall n \leq N$, one can always continue the coloring with a complete bipartite graph $K_{m,n}$ with  $(m+n)  \leq  5$. Furthermore, if the \textit{H}-coloring is odd(even), the coloring can be continued with a bipartite graph with $(m+n) = 6$ only for configurations with \textit{n} and \textit{m} odd(even) numbers.
\end{enumerate}

\subsection{Feedforward and ResNet topologies}
The existence of a homomorphism between a standard, pre-2014 feedforward connection is banal, as in their graph representation this connectivity is a
path, and there always exists a homomorphism, e.g.$\{ a_{1}, b_{1}, a_{2}, b_{2}, a_{3}, b_{3} ..., a_{n} \}$, that maps a path.

Residual networks (ResNet) are relatively deep networks that employ skip connections or shortcuts to jump over certain layers~\citep{Y2016heCCVPR}.
\begin{theorem}\label{thm:resnet}
Any ResNet architecture can be \textit{H}-colored onto the 5PP.\end{theorem}
\begin{proofsketch}
Regarding the ResNet architecture, given R2, we assimilate, without loss of generality, standard residual connections to connections from the input
memory of one vertex to the output of the next (in Fig.~\ref{Rules}\Rthree , $a_{2}$ to $a_{3}$, thus merging to the feedforward connection from
$a_{2}$ to $a_{3}$). This  requires the shortest connections possible to implement the residual connection,
merging with the feedforward connection between two adjacent vertices. It also simplifies the ResNet graph into a series-parallel graph with maximum
out-degree $d^{o}$ of the vertices equal to 2 for three adjacent vertices, which is reached for a residual path across layers with different
strides, where one resampling layer is required as shown in Fig.~\ref{Rules}\Rtwo. Note that any other implementation of the residual connections
would have yielded the same results in terms of maximum $d^{o}$ in the graph and number of adjacent vertices required.
%The \textit{H}-colorability of the resampling layer graph doesn't depend on the parity of the coloring, since it requires three adjacent vertices with maximum $d^{o}$ $ \leq$ 4, which is verified regardless of the parity by P1.
A complete mathematical argument is provided in Appendix~\ref{apx:resnet}.
\end{proofsketch}

\subsection{DenseNet topologies}
Dense connectivity as proposed by~\citep{Y2017huangCCVPR} sees a sequence of densely connected layers, all with the same channel depth, where every
node receives as input the concatenation of all its preceding layers. The DenseNet CNN comprises dense blocks connected in series.

\begin{figure}[h!]
\centering
\includegraphics[width=1\linewidth]{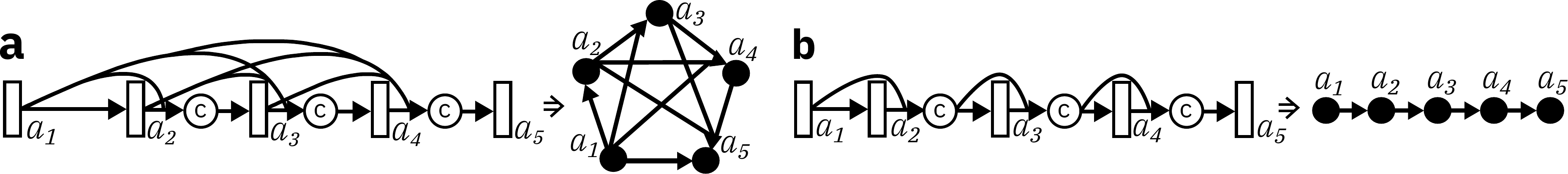}
\caption{\textbf{a.} Representation of dense connectivity as a complete graph. \textbf{b.} Representation of dense connectivity as a path graph.}
\label{densenet_graph}
\end{figure}

\begin{theorem}\label{thm:densenet}
Any DenseNet topology can be \textit{H}-colored onto the 5PP by using the maximum number of edges possible.
\end{theorem}
\begin{proofsketch}
To represent densely connected layers as a graph, there are fundamentally two possible representations, based on whether activations are
communicated to the subsequent layers before or after data aggregation implied by the concatenation operation. In the former case, in Fig.\ref{densenet_graph}\DNconn , the activations that are communicated between layers are the output activations of each layer, and \textit{n} densely
connected layers are equivalently \textit{n} vertices with an edge connecting each pair of vertices: that is, a complete graph $K_{n}$. Physically,
these edges represent channels that communicate the same number of activations. Conversely, in the latter case in Fig.~\ref{densenet_graph}\DNpath , the activations  communicated between layers are the input and output activations of each layer and, after
application of R2 and R3, in the same fashion as depicted in Fig.~\ref{Rules}\Rtwo ,\Rthree , the dense connectivity streamlines into a path
connectivity. Physically, these edges represent channels where the number of activations that are communicated increases linearly in the direction of
execution. Although representing dense blocks as paths makes their hardware mapping trivial, it makes minimal use of the entire infrastructure
and consequently channels the entire traffic on a limited number of edges, inflating the bandwidth requirement for the single communication link. As
our proposed topology is a sort of path connection of complete graphs $K_{6}$, we adopt both representations in Fig.~\ref{densenet_graph}
to distribute the communication and obtain the minimum bandwidth requirements for the physical channels. A complete mathematical argument is
provided in Appendix~\ref{apx:densenet}.
\end{proofsketch}

\subsection{Inception-style topologies}
\begin{figure}[h!]
\centering
\includegraphics[width=13.4cm]{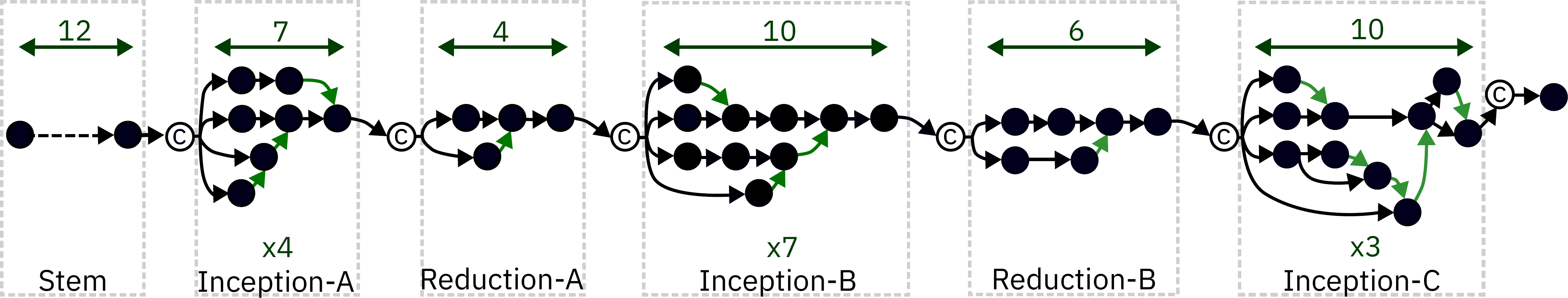}
\caption{Inception v4 architecture represented according to R1/4 featuring 17 Inception blocks connected through a series of complete bipartite
graphs $\{ K_{2,2} \textrm{(In Stem, not shown in fig.)}, K_{1,4}(\times4), K_{1,2}, K_{1,4}(\times7), K_{1,2}, K_{1,4}, K_{2,4}, K_{2,1} \}$.}
\label{inception_v4}
\end{figure}
Inception-style architectures feature a less regular, broader spectrum of connections with respect to the previous examples, comprising a sequence of
s-p graphs (``Inception blocks'') connected in series through concatenation nodes.
\begin{theorem}\label{thm:inception}
Inception v1,2,3,4 and Inception ResNet v1,2 are \textit{H}-colorable on the 5PP.
\end{theorem}
\begin{proofsketch}
Inception blocks feature a source concatenation node connecting to a maximum of four parallel branches in all Inception architectures; the number of
parallel branches always tapers in the direction of the destination concatenation node and all vertices between source and destination nodes have an
out-degree lower than or equal to 2. Note that in our graph representation of Inception networks, as discussed in design rule 4, we choose to hop all data
from layers prior to the longest latency path through the nearest temporally subsequent layer. Although this simple criterion allows implementation of
the Inception blocks of all Inception networks, data from one vertex can be hopped through any temporally subsequent vertex and can be considered a
design parameter to optimize power consumption or maximum bandwidth of the links. We can prove the
\textit{H}-colorability of the single Inception blocks with property 3 based on these patterns. A complete mathematical argument is provided in Appendix~\ref{apx:inception}.
\end{proofsketch}

\section{Quantitative Evaluation of 5PP}\label{sec:compstudy}
\begin{figure}[h!]
\centering
\includegraphics[width=14cm]{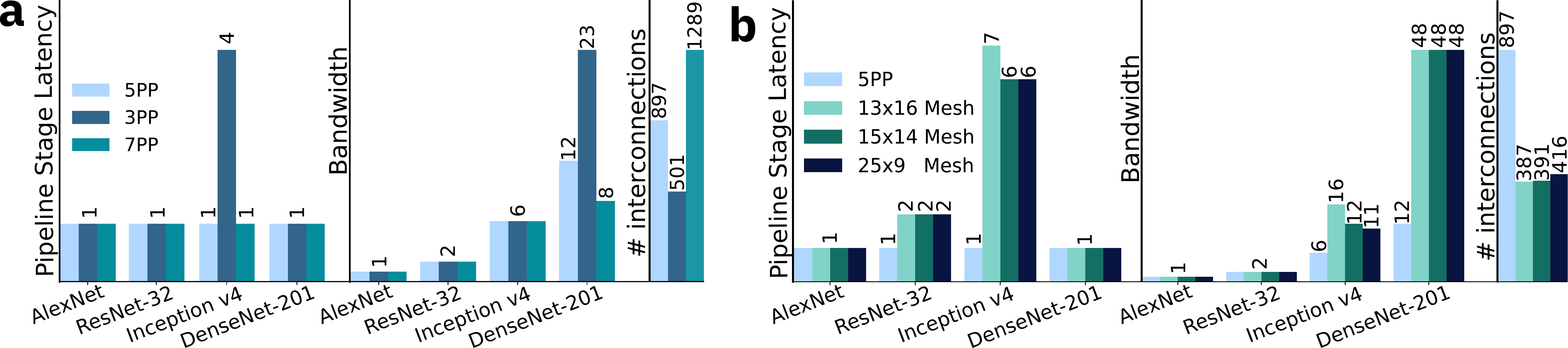}
\caption{Bar plots of pipeline stage latency and bandwidth requirements for four state-of-the art CNNs on \textbf{a.} topologies built on Algorithm~\ref{alg:5pp_construction} and \textbf{b.}  well-established topologies.} \label{fig:compstudy}
\end{figure}

This section delves into the search space defined by Algorithm~\ref{alg:5pp_construction} and explains why, among the possible complete graphs
employed as unit graphs, complete graph $K_{6}$ and the corresponding topology 5PP constitute the optimal choice. We also quantitatively compare 5PP
with well-established communication topologies.

\textbf{Evaluation metrics.} The core criterion that defines optimality is the latency of the single pipeline stage. When a homomorphism cannot be
established, the latency of the single pipeline stage would be limited by the additional computational cycles required for data movement. In Fig.~\ref{fig:compstudy}, pipeline stage latency is measured in number of computational cycles. The
bandwidth of the links is determined by the CNN layers with the greatest channel depth. We thus normalize the
bandwidth requirements by the quantity $(C_{max} \cdot bits_{act}) / T$ with $C_{max}$ maximum number of channels per layer in the network,
$bits_{act}$ the accuracy of the activations in bits and $T$ the computational cycle.

\textbf{Variations on 5PP}. Figure \ref{fig:compstudy}\PPvar{} shows a comparison of these metrics between the 5PP and two topologies built in the same fashion as the 5PP from
complete graphs $K_{4}$ (3PP) and $K_{8}$ (7PP), with the intent of showing the trend for topologies built on complete graphs with a lower and higher
number of edges. In terms of latency, 5PP never stalls the pipeline. It is not possible to establish a homomorphism for every class of networks built on $K_{n},{ }n<5$, resulting in the overall higher pipeline stage latency as is the case for 3PP. Conversely, 5PP is a subgraph of any topology
built on $K_{n},{ }n>5$, meaning the \textit{H}-coloring still holds and the pipeline is never stalled. For DenseNet, the higher the number of edges, the more effectively one can distribute the data movement, yielding lower bandwidth requirements. Indeed, distributing the communication of activations as described in Section~\ref{sec:5PPmap} yields the minimum bandwidth requirement: If communication of the activations
requires bandwidth \textit{k}, \textit{d} densely connected layers can be made to communicate synchronously on a topology built on a unit graph $K_{n}$ requiring maximum bandwidth $\left[ k +k\cdot
\Theta\left(\lceil \left( d-n \right) / (n-2)\rceil \right) \right] / T $, where $\Theta$ is the Heaviside step function and $T$ the computational cycle. These
lower requirements come at the price of instantiating multiple additional edges that have no advantage in terms of latency. In Fig.~\ref{fig:compstudy}\PPvar, the 7PP does perform 1.5$\times$ better than 5PP solely on the bandwidth requirement for DenseNet-201, at the cost of 1.44$\times$
more physical links overall.

\textbf{Existing topologies.} We now compare the performance of our topology with long-established communication topologies~\citep{Y2010sanchezACM} on the metrics defined in
Section~\ref{sec:5PP}. We consider 2D meshes with different aspect ratios  as the prior art in communication fabric topologies. Note that
mapping onto these topologies is based on the same hypotheses as  mapping onto the 5PP shown in Section~\ref{sec:5PP}. Figure~
\ref{fig:compstudy}\textbf{b}. gives the bar plot rendition of pipeline stage latency and bandwidth requirements measured in the same fashion as in
Fig.~\ref{fig:compstudy}\textbf{a}. In terms of latency, path-connected networks (AlexNet) and DenseNet in its path representation
(Fig.~\ref{densenet_graph}\DNpath) can be executed without breaking the pipeline. Conversely, this shows quantitatively how the absence of an \textit{H}-coloring between
ResNet-32 and Inception v4 impairs execution of the network by stalling the pipeline to allow communication between non-adjacent vertices. With
regard to bandwidth, thanks to the DenseNet representation described in Section~\ref{sec:5PPmap}, 5PP significantly lowers bandwidth requirements
4$\times$ with respect to the best 2D mesh performance. Although the better performance overall in 5PP costs a greater number of interconnections, their increase(2.31$\times$) is significantly lower than the  best factors of improvement in latency(7$\times$) and bandwidth(4$\times$).
\section{Case study: ResNet-32 on a CM array}\label{sec:casestudy}
\begin{figure}[h!]
\centering
\includegraphics[width=1\linewidth]{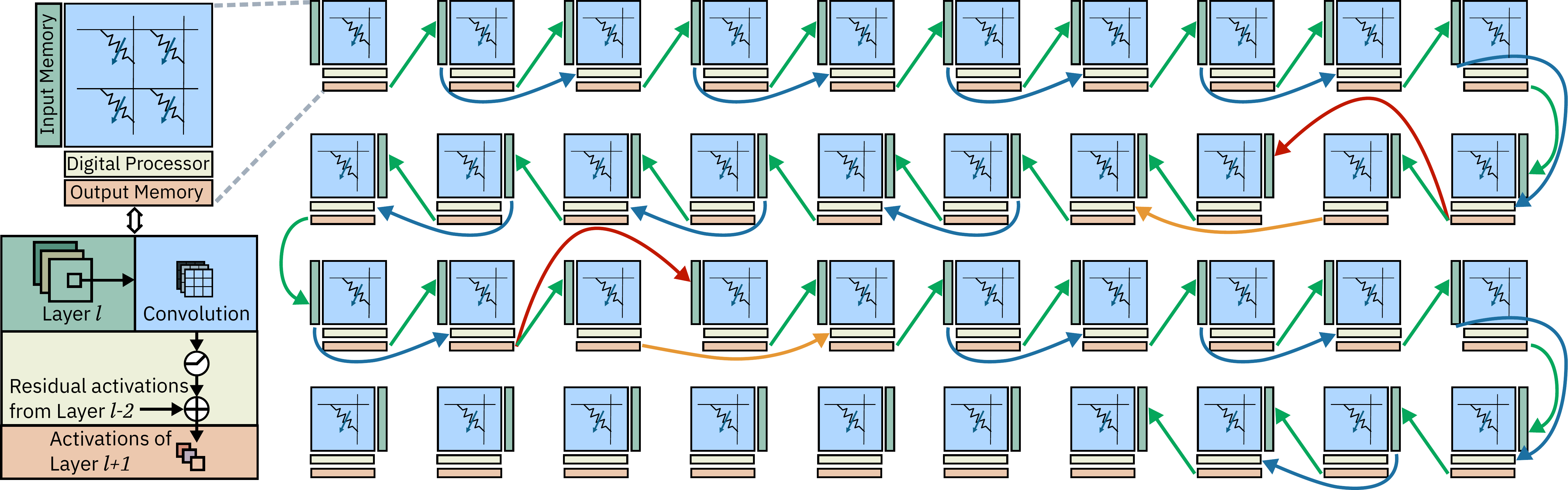}
\caption{Mapping of ResNet-32 on a 4-by-10 array of CM cores. Each layer is mapped onto one core. Arrows indicate the
communication of data between cores. Green arrows denote the propagation of activations to the adjacent layer. Blue arrows represent data
movement arising from residual connections. Orange arrows indicate re-sampled residual connections, and red arrows represent communication to
residual re-sampling layers.} 
\label{fig:rn32onCM}
\end{figure}
The physical mapping of a ResNet-32~\citep{Y2016heCCVPR} for CIFAR-10 dataset on an array of CM cores is depicted in Fig.~
\ref{fig:rn32onCM}. The network features three layer levels with channel depths equal to 16, 32, and 64, whereas the input image size is 32$\times$32
pixels. The CM cores are represented as boxes with their associated input (left) and output (bottom) memory. Each core also comprises modest digital
processing capability to scale the crossbar outputs, e.g.~to apply the batch normalization, to implement the activation functions and to
perform the residual addition. We assume 576$\times$576 for the memristive crossbar size per CM core, meaning the vector-matrix multiplication with a
matrix of size 576$\times$576 can be performed in one computational cycle, which is assumed to be \unit[100]~{ns}. During execution, the memristive
elements of the crossbar hold one convolutional weight each, whereas the input memory will store the pixel neighborhood required for the convolution.
The results of the convolution, that is a single activation across the entire channel depth, is stored in the output memory. We assume 8-bit
precision for activations, which has proved sufficient to achieve state-of-the-art accuracy for the CIFAR-10 dataset~\citep{Y2016shafieeACM}. 

As mentioned in Section~\ref{sec:cnngraphs}, the communication channels can communicate indistinctly to both the input memory (to communicate the
standard feedforward activations) or to the output memory (to communicate the activations used for residual addition).  The dataflow occurs row-wise
starting from the core located at the top left. Whenever one set of activations is computed by a core, it is communicated to the core assigned to
compute the subsequent layer and stored in its input memory. Computation on one core begins when it has received sufficient activations to perform
the convolution operation corresponding to its own assigned layer. The dataflow and network/dataset specifications define the memory and bandwidth
requirements. The memory requirement is defined by the minimum number of activations to be stored per feature map. The proposed topology
ensures that the pipeline is never stalled irrespective of the bandwidth. However, the most efficient implementation would be having  
sufficient bandwidth for all data transfer to occur in parallel with the computation cycle. Conversely, if the bandwidth is not sufficient, a constant
communication overhead must be added to each computational cycle. In the former case, we estimate the required bandwidth per channel to be
approximately \unit[5]~{Gbps}, which is state-of-the-art for on-chip links~\citep{Y2017saccoVLSI}. Note that the link that delivers activations from both the feedforward and residual paths would be physically
implemented as two separate channels.

\section{Conclusions}
We introduced 5 Parallel Prism (5PP), an interconnection topology for executing CNNs on an array of CM cores. We then proved the executability of ResNet, Inception, and DenseNet networks on the proposed communication fabric by proving the existence of a homomorphism between a consolidated graph representation of the CNNs and 5PP. Moreover, we validated the efficacy of our approach by comparing the proposed topology to various 2D meshes on the metrics of inference latency as well as bandwidth requirements per communication channel. Finally, we provided a case study with the physical mapping of ResNet-32 on an array
of CM cores. The presented work is a significant step towards developing general-purpose DNN accelerators based on in-memory computing.
\medskip
\bibliography{5PP}

\clearpage
\appendix
\section{Homomorphism Proofs}

\subsection{Proof of Theorem \ref{thm:resnet}}\label{apx:resnet}
Let a graph represent a ResNet topology according to the rules in Section~\ref{sec:cnngraphs}. Such a graph is a series connection of paths (R3,
Fig.~\ref{Rules}\Rthree) and resampling layers (R2, Fig.~\ref{Rules}\Rtwo) in that order. We prove a homomorphism by \textit{H}-coloring the two
elements in sequence. Any path can be iteratively \textit{H}-colored in  5PP as proved above, ending with a certain parity. Then, the resampling layer
is \textit{H}-colored, which is equivalent to coloring a complete graph $ K_{3} $, represented in Fig.~\ref{Rules}\Rtwo { }  by
vertices $ a_{3}$,  $a_{4}$ and $a_{5}$. Regardless of the parity of the coloring before the resampling layer, according to P1 there are always three
uncolored vertices on which to color one $ K_{3}$; there are also always enough edges, as $ K_{3} $ is always a subgraph of $ K_{6} $  for any three
vertices belonging to  $ K_{6} $.

\subsection{Proof of Theorem \ref{thm:densenet}}\label{apx:densenet}
Let a graph represent an \textit{n}-layer DenseNet topology. Assume there exists a homomorphism that maps its layers $\lbrace v_{1}, ..., v_{n}
\rbrace$ to a sequence $\lbrace a_{1}, b_{1}, a_{2}, b_{2}, ..., b_{m} \rbrace$ on 5PP using all edges, ordered as in Fig.~\ref{fig:complete_prism}. If the number of densely connected vertices is $\leq 6$, they all belong to the same $K_{6}$ $\lbrace a_{1}, b_{1}, ...,
a_{3} \rbrace$, and the \textit{H}-coloring follows from Fig.~\ref{densenet_graph}\DNconn . If the number of densely connected layers is $> 6$, the complete
connection of all individual instances of $K_{6}$ is used, but there are vertices that do not belong to the same $K_{6}$. For example for eight layers, vertices
$b_{4}$ and $a_{4}$ are connected through a complete graph to $\lbrace a_{2}, ..., b_{3} \rbrace$, but not to $a_{1}$ and $b_{1}$. The communication
of activations from  $a_{1}$ and $b_{1}$ to vertices $b_{4}$ and $a_{4}$ is thus distributed among the communication from the four vertices $\lbrace
a_{2}, ..., b_{3} \rbrace$ belonging to the same $K_{6}$ as $b_{4}$ and $a_{4}$ as described in Fig.~\ref{densenet_graph}\DNpath . In
general, for an arbitrary number \textit{n} of densely connected layers, the \textit{i-th} vertex, where $i$ is an even(odd) number $>6 $ and able to communicate
with the previous 5(4) vertices through the communication fabric. The edges between these 5(4) vertices and $v_{i}$ communicate the output activations
of the previous 5(4) layers, plus the activations from vertices $\lbrace v_{1}, ..., v_{i-6(5)} \rbrace$.

\subsection{Proof of Theorem \ref{thm:inception}}\label{apx:inception}
Let a graph represent one Inception topology. Inception topologies are made up of Inception blocks connected by concatenation nodes. Inception blocks
are s-p graphs with four or less parallel branches and vertices with $d^{o}$ of at most 2. We first prove the coloring of the individual Inception
blocks. From P1, every vertex in a 5PP belongs to at least one $K_{6}$. Given a maximum of four parallel branches, they are always colorable on a
plane with vertices $\{ a_{n}, b_{n}, a_{n+1}, b_{n+1} \}$, all belonging to a single $K_{6}$ with vertices $\{ a_{n}, b_{n}, a_{n+1}, b_{n+1},
a_{n+2}, b_{n+2}  \}$. As $\{ a_{n}, b_{n}, a_{n+1}, b_{n+1} \}$ can be chosen belonging to the same $K_{6}$, every vertex will have possible $d^{o}$
equal at least to the number of uncolored vertices in that $K_{6}$, which is equal to 2. What remains to be proven is the colorability of the
concatenation nodes through which they are connected. From P3, complete bipartite graphs with $m+n<5$ are always colorable. This condition applies to
all concatenations in Inception architectures, which are therefore mappable.
\end{document}